\documentclass[letterpaper, 10 pt, conference]{ieeeconf}  % Comment this line out if you need a4paper

\IEEEoverridecommandlockouts                              % This command is only needed if 
\title{\LARGE \bf
Motion Planning through Demonstration to Deal with\\ Complex Motions in Assembly Process 
}

\author{Yan Wang$^{1}$, Kensuke Harada$^{1}$$^{2}$ and Weiwei Wan$^{1}$$^{2}$% <-this % stops a space
\thanks{$^{1}$Graduate School of Engineering Science,
        Osaka University, Japan}%
\thanks{$^{2}$National Inst. of AIST, Japan. 
        *Correspondent author: Yan Wang:
        {\tt\small yan@hlab.sys.es.osaka-u.ac.jp}}%
}
\usepackage[keeplastbox]{flushend}
\usepackage{graphicx}
\graphicspath{ {./pics/} }
\usepackage[bottom]{footmisc}
\usepackage[utf8]{inputenc}
\usepackage{multirow}
\usepackage{multicol}
\usepackage{tabularx}

\begin{document}
\maketitle
\thispagestyle{empty}
\pagestyle{empty}

%%%%%%%%%%%%%%%%%%%%%%%%%%%%%%%%%%%%%%%%%%%%%%%%%%%%%%%%%%%%%%%%%%%%%%%%%%%%%%%%
\begin{abstract}

Complex and skillful motions in actual assembly process are challenging for the robot to generate with existing motion planning approaches, because some key poses during the human assembly can be too skillful for the robot to realize automatically. In order to deal with this problem, this paper develops a motion planning method using skillful motions from demonstration, which can be applied to complete robotic assembly process including complex and skillful motions. In order to demonstrate conveniently without redundant third-party devices, we attach augmented reality (AR) markers to the manipulated object to track and capture poses of the object during the human assembly process, which are employed as key poses to execute motion planning by the planner. Derivative of every key pose serves as criterion to determine the priority of use of key poses in order to accelerate the motion planning. The effectiveness of the presented method is verified through some numerical examples and actual robot experiments.

\end{abstract}

%%%%%%%%%%%%%%%%%%%%%%%%%%%%%%%%%%%%%%%%%%%%%%%%%%%%%%%%%%%%%%%%%%%%%%%%%%%%%%%%
\section{Introduction}

\begin{figure}[b]
      \centering
      \includegraphics[width=8.5cm]{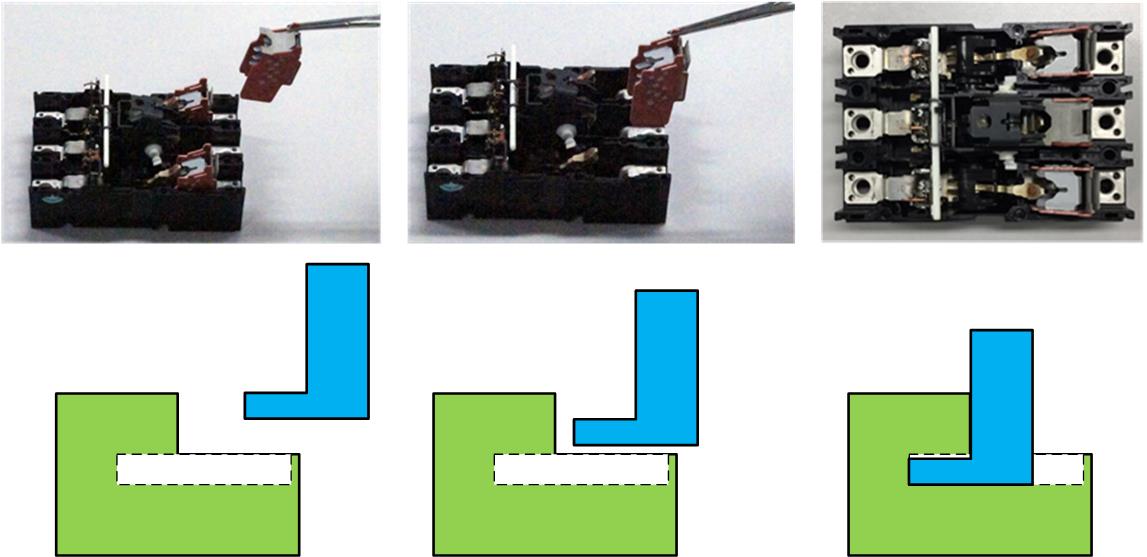}
      \caption{The assembly process of a condenser and its simplified model by using an L-shaped object insertion.}
      \label{fig:condenser}
   \end{figure}

\begin{figure}[t]
      \centering
      \includegraphics[width=3cm]{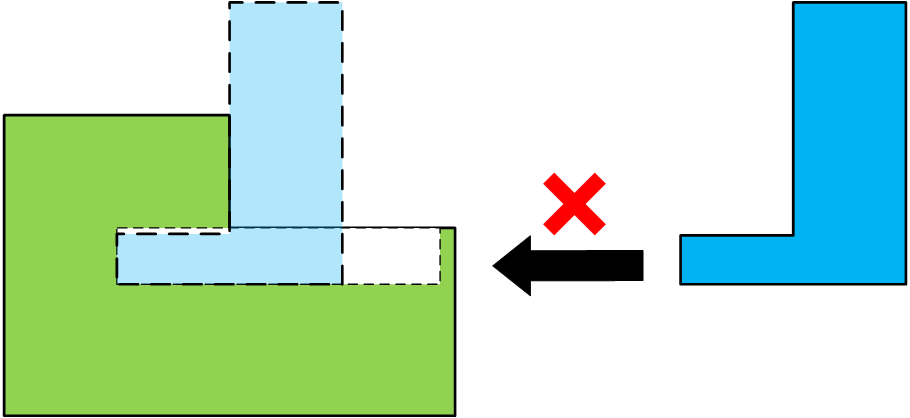}
      (a)
      \includegraphics[width=1.5cm]{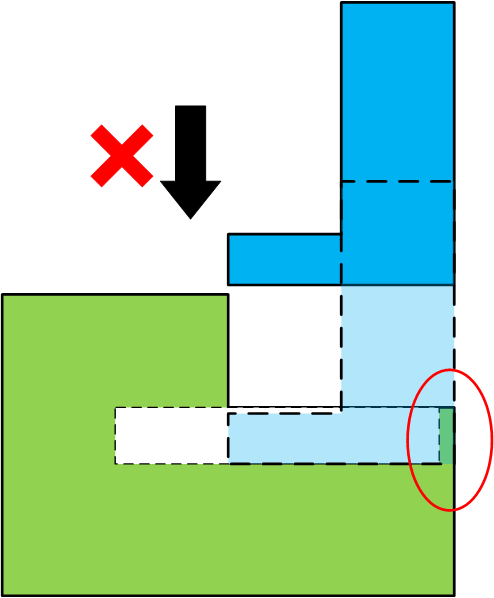}
      (b)
      \caption{Difficulties in the \textbf{L} insertion. (a) \textbf{G} is closed on both ends so that \textbf{L} cannot be inserted into the hole from one end directly; (b) The length of \textbf{G} exposed outside is a bit shorter than the bottom length of \textbf{L}, which makes it impossible to put \textbf{L} into \textbf{G} with its bottom horizontal and insert it into the hole.}
      \label{fig:insert examples}
   \end{figure}

\begin{figure}[b]
       \centering
       \includegraphics[scale=0.23]{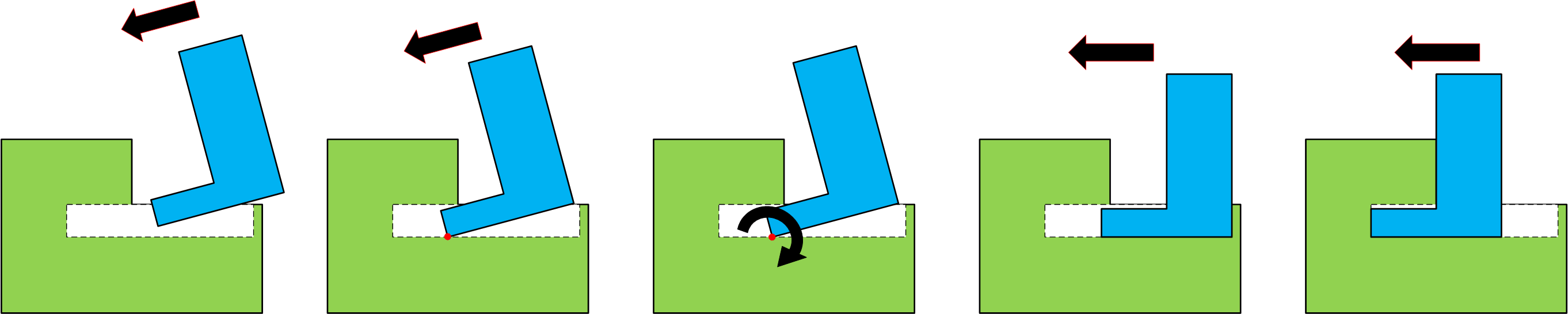}
       \caption{Human motions to insert \textbf{L} into the hole of \textbf{G}. 
       1) Tilting properly and translating \textbf{L} towards the groove until one side of its bottom enters the hole and touches the groove; 2) Rotating \textbf{L} around the contact point, \textbf{A}, to make its bottom contact the groove completely (slide often occurs); 3) Translating and inserting \textbf{L} into the hole completely.}
       \label{fig:insert human}
   \end{figure}

In factory environment, industrial robots are expected to finish the product assembly process automatically. However, the robotic assembly process can include very skillful and complicated motions, for which the solution of motion planning is very difficult to find despite that it actually exists.
For example, the assembly process outline of a condenser is shown in Fig.\ref{fig:condenser}, and it can be simplified into an L-shaped object insertion problem in narrow space. There is an L-shaped object named \textbf{L}, and an object with a groove named \textbf{G}. The target is to insert \textbf{L} into the hole of \textbf{G}. However, there are some factors making the \textbf{L} insertion hard to execute as Fig.\ref{fig:insert examples} shows. Therefore, a worker has to act skillful motions to finish this assembly process according to steps in Fig.\ref{fig:insert human}.
   
Traditional motion planning can hardly deal with such skillful and complex motion planning because the motions can be too delicate to generate a proper path automatically, despite that it can generate collision-free path. On the other hand, these skillful and complex motions are pretty common in human assembly process, but collision often happens due to the flutter of hands. Therefore, we present an idea of guiding the motion planning with the motion capture data from human demonstration in a robotic motion planner, which enables the robot to find the solution path of these skillful motions with the least collision probability.

In order to make it explicit what human does to the object manipulated during an assembly process, we use augmented reality (AR) markers to capture some key poses of the object manipulated by human during an assembly process. Because violent changes are usually main reasons of the failure of motion planning, we rearrange the key poses in descending order of their derivatives in the motion trajectory so that the most important pose can be inserted into the planner in priority, enhancing the speed of motion planning. 

In this research, grasp configurations are generated by the planner automatically and assembly positions/orientations are determined in advance. Besides, since we just concentrate on skillful motions in single task during a complete assembly process, assembly sequence is not discussed, either. 

The rest of this paper is organized as follows: First, we introduce some related work of our research in Section II. Then, we outline the proposed approach with an example in Section III and give a detailed description of the methodology in Section IV. In Section V, numerical simulations and actual experiments are implemented to validate the effectiveness of the presented method. Finally, we summarize our work and discuss the future work in Section VI.
    
\section{Related Work}

In this section, we review the related studies of our work in three aspects: motion planning, robot teaching by demonstration, and visual recognition and tracking.

\subsection{Motion Planning}

Robot motion planning (RMP) generates robotic motion of manipulating parts to achieve tasks. \cite{koga1992experiments} was the groundbreaking work comparing the joint space and workspace approaches. They also presented an approach to compute the collision-free trajectories for multi-arm planning problem \cite{koga1994planning}. 

After that, some typical RMP algorithms appeared, including the probabilistic roadmaps approaches to search collision-free motion in the joint space \cite{simeon2004manipulation} \cite{kavraki1994probabilistic}, and the rapidly-exploring random trees(RRT) \cite{lavalle2000rapidly}. As for the recent work like \cite{berenson2012robot} \cite{phillips2015speeding}, researchers made use of historic data to improve the system performance.

Compared to conventional motion planning methods, our method improves the planning effectiveness to certain extent via utilization of the human expertise.

\subsection{Robot Teaching by Demonstration}

“Teaching by demonstration” is a method of generating a robot program that makes a robot do  the same task as the task that a human operator demonstrates \cite{onda2002teaching}. Robot teaching by demonstration requires the acquisition of example trajectories, which can be captured in various ways \cite{zhu2018robot}, mainly including kinesthetic demonstration, motion-sensor demonstration, and teleoperated demonstration. In kinesthetic demonstration, robotic hands are guided by a human demonstrator and the movements are recorded directly on the learning robot \cite{calinon2007learning} \cite{ye2016guided}. For motion-sensor demonstration, marker-based tracking devices are deployed because of the high accuracy compared to computer vision \cite{ude2004programming} \cite{ruchanurucks2006humanoid}. Teleoperated demonstration can establish an efficient communication and operation strategy between humans and robots, which has been applied in performing assembly tasks \cite{chen2003programing} \cite{bohren2013pilot}.

Without redundant third-party devices, this paper makes use of an AR marker tracking system to capture poses of objects during an assembly process. These poses serve as key poses for the planner to generate motion sequence to complete the assembly task. 

\subsection{Visual recognition and tracking}

In robotic manipulation, visual recognition and tracking technique is rather critical, which has been well studied in the last decades. Researchers has developed various kinds of visual recognition and tracking approaches, such as determining the poses of objects from 2D images \cite{david2004softposit}, using color co-occurrence histograms and geometric modeling to estimate the 6-DOF pose of objects \cite{ekvall2005object}, using shape/contour to represent objects \cite{lin2007hierarchical} \cite{gavrila2000pedestrian}, and 3D matching with data from stereo cameras \cite{rusu2010fast} and structured light \cite{moeslund2007pose}. In recent years, the augmented reality (AR) markers are widely used to obtain pose information \cite{babinec2014visual} \cite{kam2018improvement}, which possess the advantages of easier recognition and higher precision \cite{garrido2014automatic}.

In this paper, we stick AR markers to the manipulated object in order to detect and track it during the assembly process. With the AR markers, the poses of the object can be recognized and recorded quickly and precisely.

\section{Overview of the Approach}

In certain assembly process requiring complex and skillful motions, it can be challenging for the robot to generate proper motions to complete a task. To promote clarity, we still present our approach with the \textbf{L} insertion in Section I as an example. 

It cannot be denied that the solution of the \textbf{L} insertion motion planning problem exists. However, it is very difficult to find. Thus, we consider guiding the solution path by using the motion capture data, i.e. the human demonstration poses. In the motion planner, the inputs are defined as 

1) The mesh models of \textbf{L} and \textbf{G}; 2) The initial and final poses of \textbf{L}; 3) Key poses of \textbf{L} captured from human demonstration.

\begin{figure}[t]
      \centering
      \includegraphics[scale=0.24]{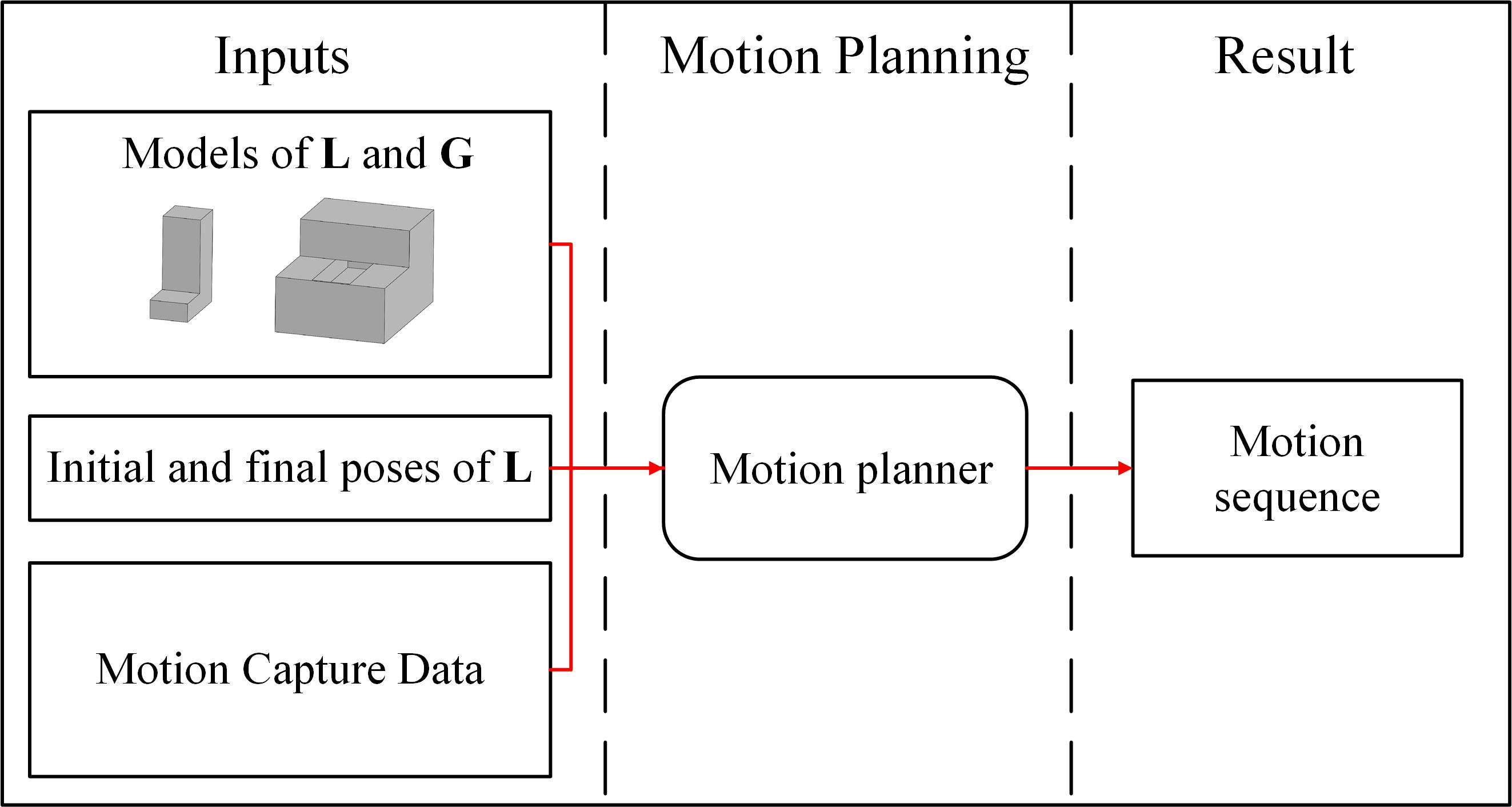}
      \caption{ Overview of the presented approach. It takes models of \textbf{L} and \textbf{G}, initial and final poses of \textbf{L}, and key poses of \textbf{L} obtained by AR marker as inputs of the motion planning, whose result is a proper motion sequence to complete the assembly process.}
      \label{fig:workflow}
  \end{figure}

Especially, for the demonstration, a real-time and precise approach is demanded to recognize and track \textbf{L} through the process. For convenience, we utilize the AR marker to achieve this purpose and the details will be discussed in Section IV. Given these inputs, the planner then deal with the data and generate an appropriate motion sequence to finish the \textbf{L} insertion process successfully. The general workflow of our approach is shown in Fig.\ref{fig:workflow}

\section{Method}

In this section, we present our method to implement motion planning using demonstration poses.

\subsection{Demonstration}

To obtain poses of actual manipulated objects, a vision system is demanded to track and capture them during the assembly process. In this paper, we capture the poses of objects with AR markers.

% Fig.\ref{fig:objwithmarker} shows an example of objects mentioned in Section III with AR markers. Especially, for \textbf{L}, a marker board bigger than the area of its cross section is attached in order to acquire better precision.
During the assembly process, \textbf{G} is generally fixed to the table and cameras in the vision system track \textbf{L} and capture its key poses. Actually, the vision system simultaneously captures absolute poses of both \textbf{L} and \textbf{G} in the world coordinate system and then calculates the relative pose of \textbf{L} to \textbf{G}.

We suppose that the pose of the fixed object \textbf{G} is $$^p\textbf{T}_G = [^p\textbf{R}_G,\ ^p\textbf{P}_G]$$ and \textbf{L} is given an initial pose $$(^p\textbf{T}_L)_0 = [(^p\textbf{R}_L)_0,\ (^p\textbf{P}_L)_0]$$ in the coordinate system of the virtual environment.
After the vision system calculating the relative poses of \textbf{L} to \textbf{G}, which are $$(^G\textbf{T}_L)_n=[(^G\textbf{R}_L)_n,\ (^G\textbf{P}_L)_n], n = 1, 2, 3 ...$$ respectively on the inserted \textit{n} key points chosen from the demonstration key poses, the planner utilizes the result to determine absolute poses of \textbf{L} in the world coordinate system of the planner virtual environment through $$(^p\textbf{T}_L)_n =\ ^p\textbf{T}_G * (^G\textbf{T}_L)_n, n = 1, 2, 3...$$
    
Then, the planner generates a motion path to achieve the assembly process with these key poses relative to the world coordinate system.

    % \begin{figure}[t]
    %     \centering
    %     \includegraphics[scale=0.12]{pics/LandGwithmarker.jpg}
    %     \caption{\textbf{L} and \textbf{G} with AR markers}
    %     \label{fig:objwithmarker}
    % \end{figure}
    
\subsection{Motion Planning}

In conventional motion planning, a motion sequence is generated automatically without human intervention. However, in skillful assembly process, such as the narrow-space insertion problem above, it is obvious that finding a collision-free path to the target completely automatically can be rather difficult and time consuming. On this circumstance, human demonstration is introduced to assist the motion planning. 

The strategy is that when it fails to generate a collision-free motion path or the planning time is more than a threshold $t_e$, the planner takes a new key pose to execute another planning process. This workflow loops until the motion path is found and the time cost is relatively short.

The motion planning algorithm adopted by the planner is the DDRRT-Connect, a simple and efficient randomized algorithm for solving single-query path planning problems in high-dimensional configuration spaces \cite{kuffner2000rrt}. In robotic assembly process, not only the collision between manipulated object and obstacles is considered, but also the collision between the robot and obstacles around. 

\begin{figure}[b]
                \centering
                \includegraphics[width=5cm]{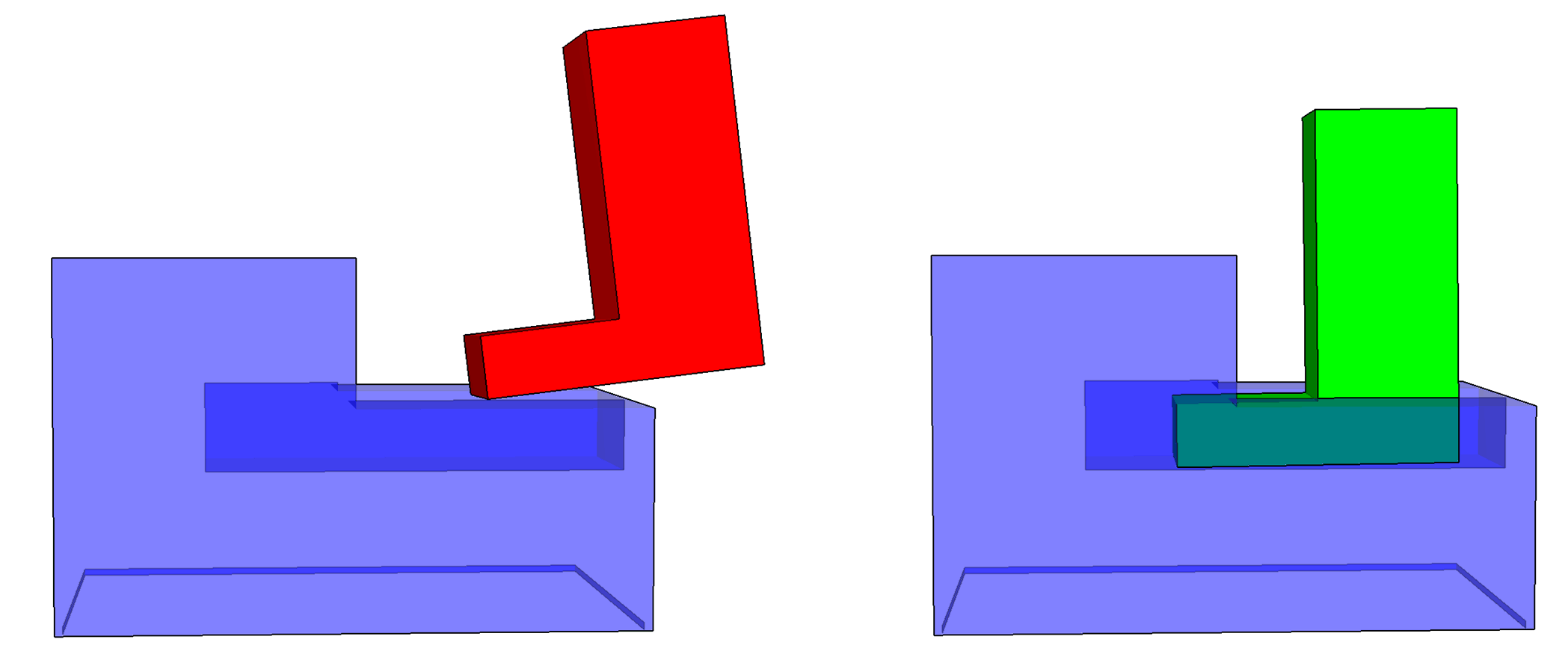}
                \caption{The initial pose $P^I$ (left) and the final pose $P^F$ (right)}
                \label{fig:initfinpose}
            \end{figure}

To begin with, we discuss the collision-free motion planning of the manipulated object. In order to enhance precision, collision mesh is utilized as the collision model. The collision detection is implemented through testing collision pairs with the collision detection library in \textit{Panda3D}. There are 6 parameters to decide the pose of the manipulated object, \textbf{L}, in 3D configuration space, \textit{x}, \textit{y}, \textit{z}, \textit{R} (roll), \textit{P} (pitch), and \textit{Y} (yaw). The initial pose $P^I$ and final pose $P^F$ of the motion planning are set in advance as Fig.\ref{fig:initfinpose} shows. 
It should be emphasized that they are different from the actual poses in practice, only chosen for simplifying the motion planning process, because the rest of the motions are relatively uncomplicated to plan. Given $P^I$ and $P^F$, the planner manages to find a collision-free motion path between them avoiding colliding with the obstacle \textbf{G}. Such kind of path is usually challenging and time-resuming to find merely with the traditional motion planning algorithm because of the narrow manipulation space. Thus, human demonstration poses are involved to assist the motion planning in finding the proper path, where the order of inserting human demonstration poses becomes an issue to discuss.

\begin{figure}[t]
                    \centering
                    \includegraphics[width=8.65cm]{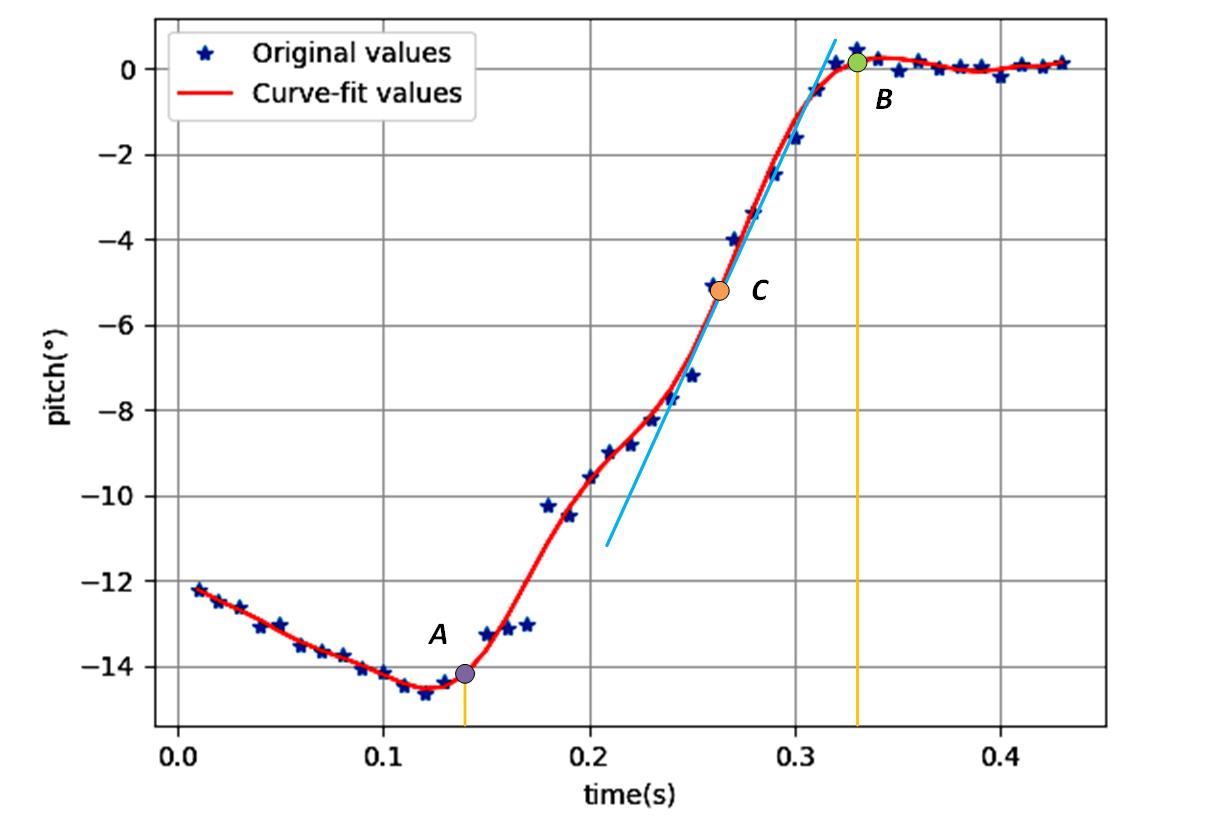}
                    \caption{The demonstration poses and the fitting curve. The part from \textit{\textbf{A}} to \textit{\textbf{B} }is the most important area where we extract key poses, The key poses are inserted in descending order of their derivative and \textit{\textbf{C}} is the first one. }
                    \label{fig:pitch}
                \end{figure}

There can be many key poses captured in a whole motion trajectory, with different weights of importance. An ideal condition is to finish the motion planning with as few as possible demonstration poses so that the motion planning takes less time. Therefore, vital poses are expected to insert in priority. Generally, the most decisive poses are often with the highest rate of change, which is the main reason of planning failure. In the \textbf{L} insertion, the most difficult part is the step 2) mentioned in Section III and \textit{P} is the parameter varying the most violently in this period. Therefore, we execute polynomial curve fitting on \textit{P} of demonstration poses as Fig.\ref{fig:pitch} shows and compute the derivative $D_n$ on all the n key poses of the obtained motion trajectory with the \textit{scipy} library. Then, we rank these poses in descending order of $D_n$ and set them as candidate poses to insert into the planner in order later. We can see that the part from \textit{\textbf{A}} to \textit{\textbf{B}} is the most important area where we extract key poses, and among them \textit{\textbf{C}} possesses the highest change rate so that it should be the first pose to insert.

\begin{figure}[t]
                    \centering
                    \includegraphics[width=8.4cm]{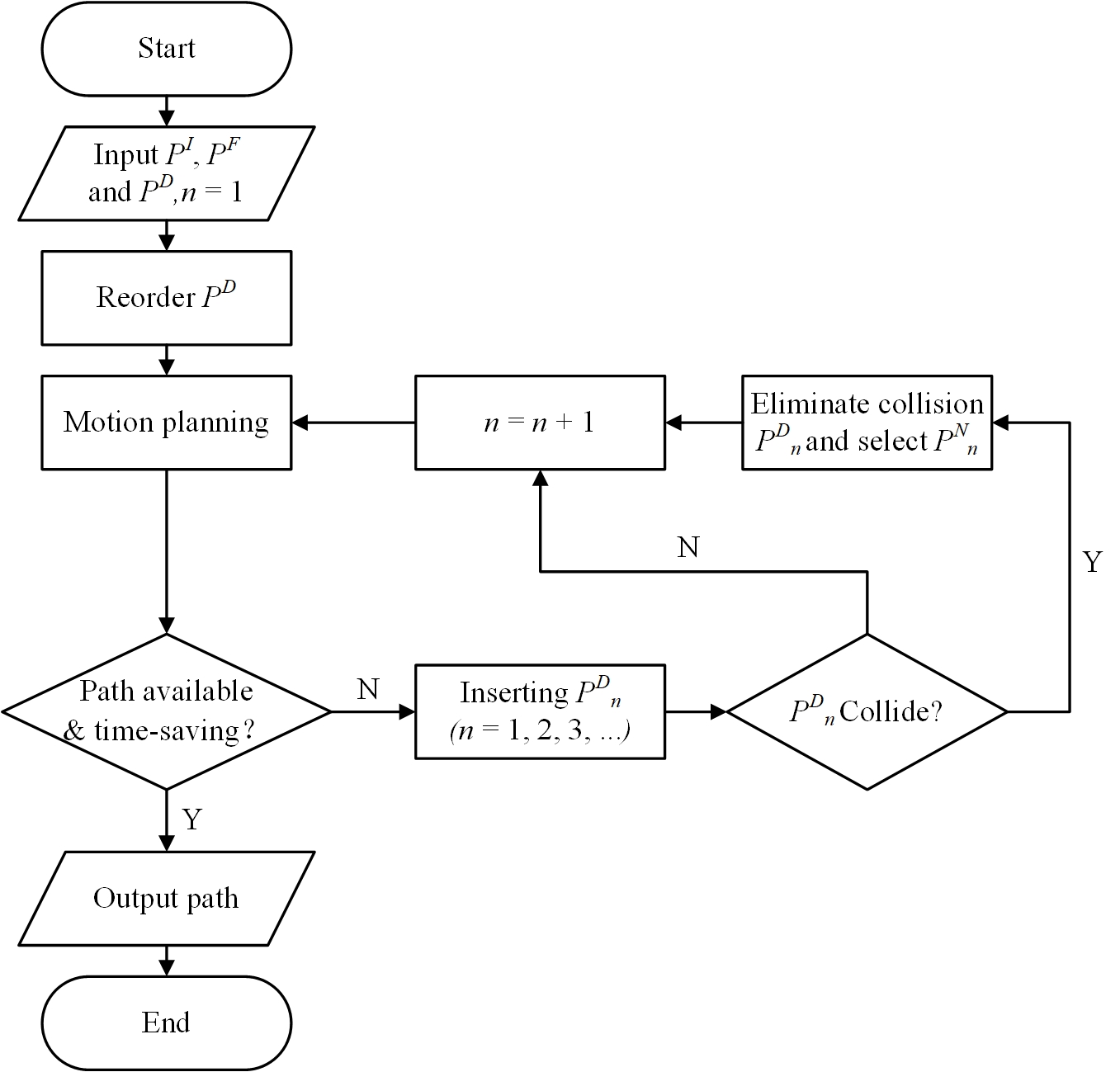}
                    \caption{The detailed workflow of motion planning with demonstration data. Demonstration poses where collision occurs are eliminated and replaced by their nearest collision-free random poses.}
                    \label{fig:mpflow}
                \end{figure}

Besides, collision regularly occurs in human demonstration so that it is necessary to deal with captured human demonstration poses where collision is detected in the planning process. After inserting the $n^{th}$ demonstration pose $P^D_n (n\ge1)$ into the planner, the planner execute motion planning between $P^D_n$ and $P^I (n=1)$ or the previous demonstration pose $P^D_{n-1} (n\ge2)$. If collision occurs at $P^D_n$, the planner will eliminate $P^D_n$ from the planner and set the nearest collision-free random pose $P^N_n$ as the new start pose until reaching the final pose in relatively short time. $P^N_n$ is randomly generated by the DDRRT-Connect algorithm. Through the method above, a collision-free motion path of \textbf{L} taking only collision between \textbf{L} and \textbf{G} into consideration is obtained. Fig.\ref{fig:mpflow} shows the entire workflow of this method.
                
Then, the collision between the robot and obstacles is involved, including the arm-obstacle collision and gripper-obstacle collision specifically. In our research, the configuration space of the robot is large enough so that the arm can be guaranteed to be collision-free. Making the issue conciser, the grasp configuration is determined in advance as a collision-free one and thus the relative pose of the object and the gripper is acquired. Therefore, sufficient collision-free poses of the gripper tip are known and a sequence of parameters can be obtained through solving the inverse kinematics. The parameters are the waist rotation angle $\theta_{waist}$, and the \textit{m} arm joint rotation angles from $\theta_1$ to $\theta_m$. Eventually, through the DDRRT-Connect motion planning, a motion path of joint angles leading to the corresponding collision-free path of manipulated object can be obtained. In the following section, successful experiments using this method and corresponding settings will be presented.

\section{Experiments and Analysis}
In this section, we propose experiments and analysis of our research. In order to verify the proposed approach, we implemented both simulation and real-world experiments. The computational platform used in our experiments was a PC with Intel Core i5-8250U CPU and 24.00GB memory. The programming language was Python 3 and the software platform was PyManipulator\footnote{https://gitlab.com/wanweiwei07/wrs}. As for the vision system, USB 3.0 web cameras with 1920*1080 resolution were used. Finally, the Nextage OPEN was employed in the actual robot experiment.

\subsection{Demonstration and Simulation}
\begin{figure}[b]
        \centering
        \includegraphics[width=8cm]{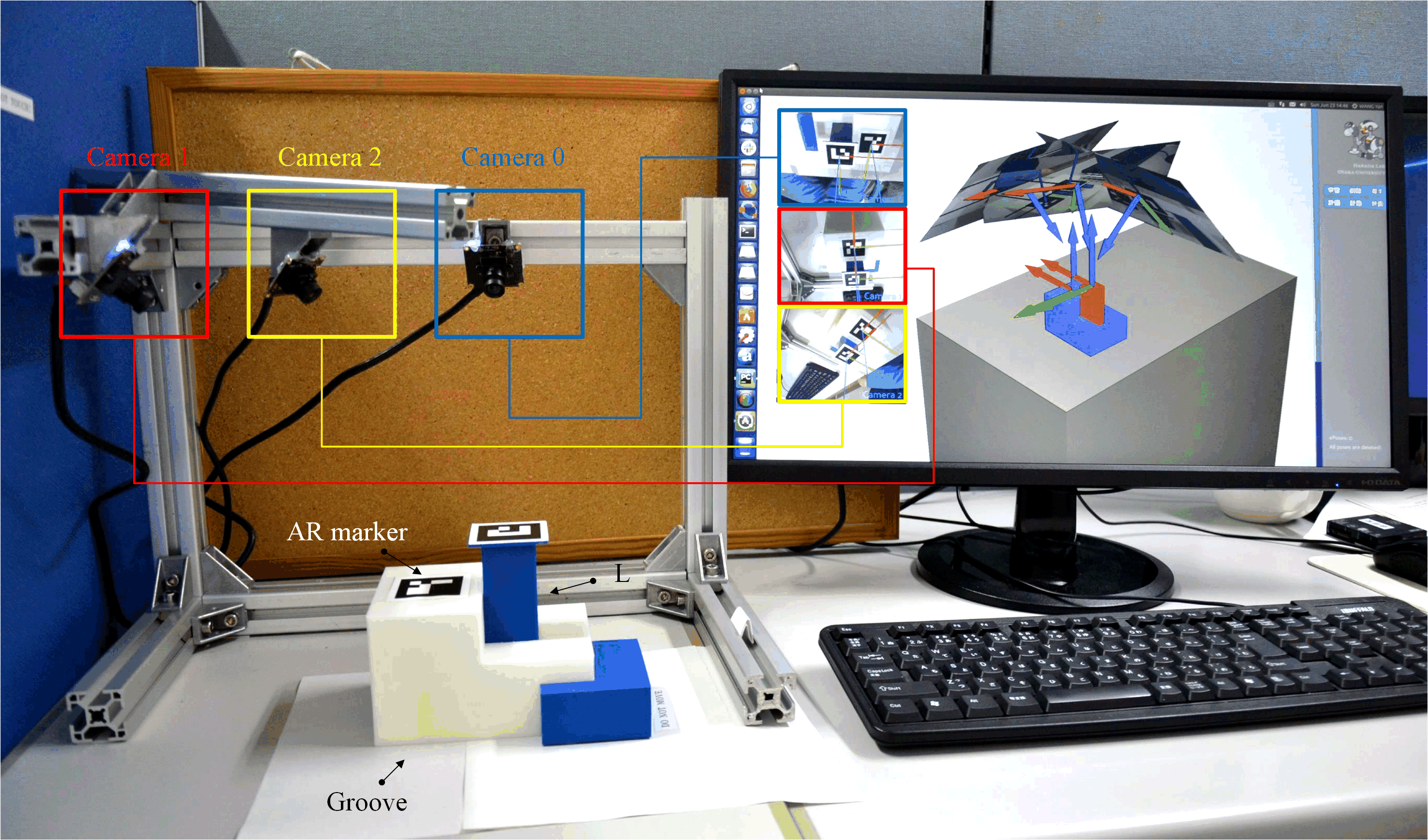}
        \caption{Vision system}
        \label{fig:visionsystem}
    \end{figure}

The vision system constructed in the experiment is shown in Fig.\ref{fig:visionsystem}.
    
\begin{figure*}[t]
            \centering
            \includegraphics[width=17cm]{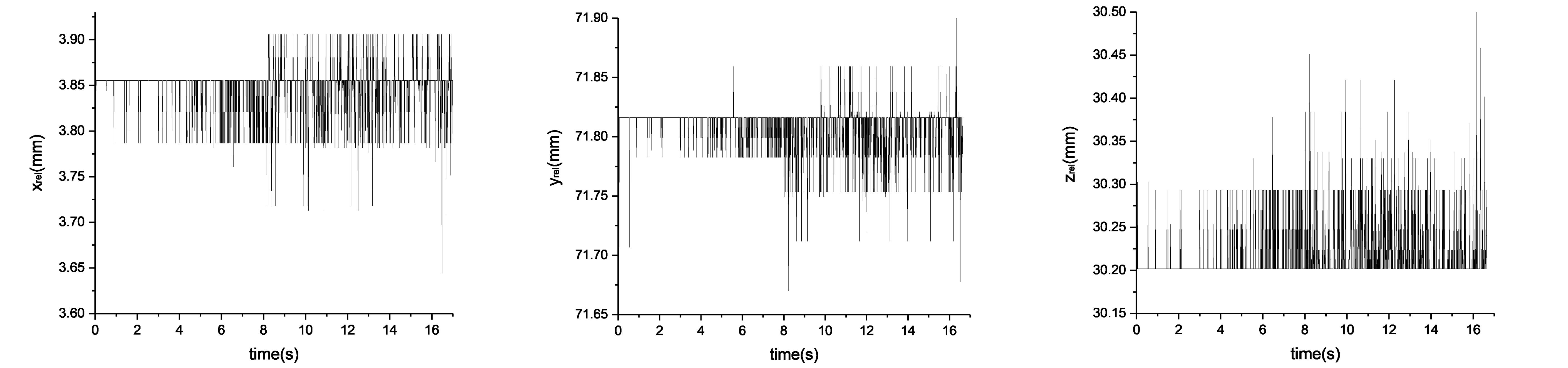}
            (a)
            \includegraphics[width=17cm]{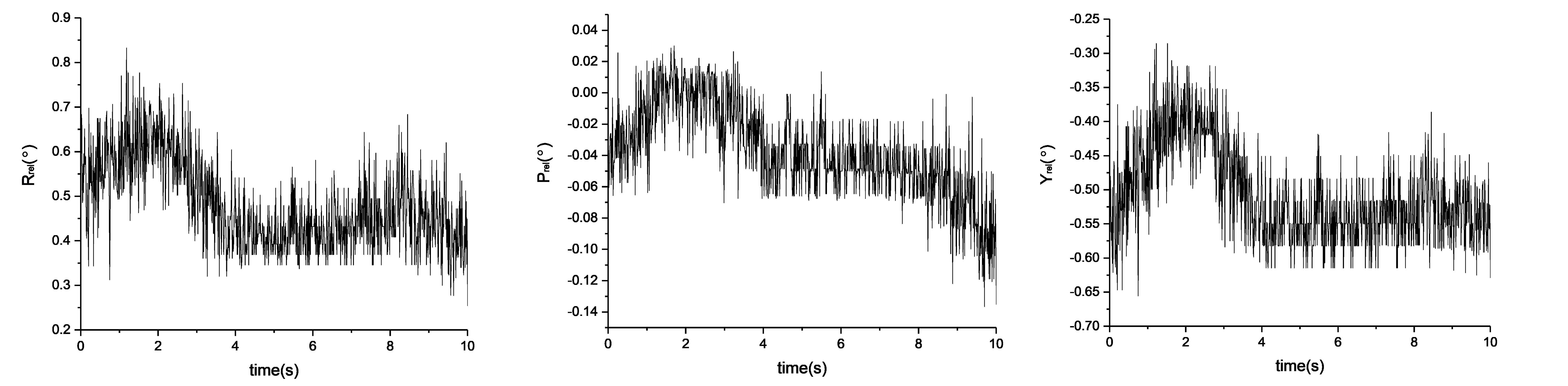}
            (b)
            \caption{Relative position and pose of \textbf{L} and \textbf{G} in the world coordinate system.}
            \label{fig:xyzrpyrel}
        \end{figure*}

It consists of a manipulation platform, three USB 3.0 cameras fixed on a frame whose IDs are 0, 1, and 2, and a monitor. The system detects AR markers on objects appearing in the vision field of the three cameras, tracks them and displays corresponding virtual models on the monitor in real time. For verifying the stability and precision of the demonstration using AR marker, graphs describing the relative position and pose of virtual models of \textbf{L} and \textbf{G} in the world coordinate system is given in Fig. \ref{fig:xyzrpyrel}. It can be concluded that the magnitude of the deviation is rather low and the relative position is almost the same to the real objects, which indicates that the vision system is stable and precision enough for the human demonstration.

 \begin{figure*}[h]
            \centering
            \includegraphics[width=17cm]{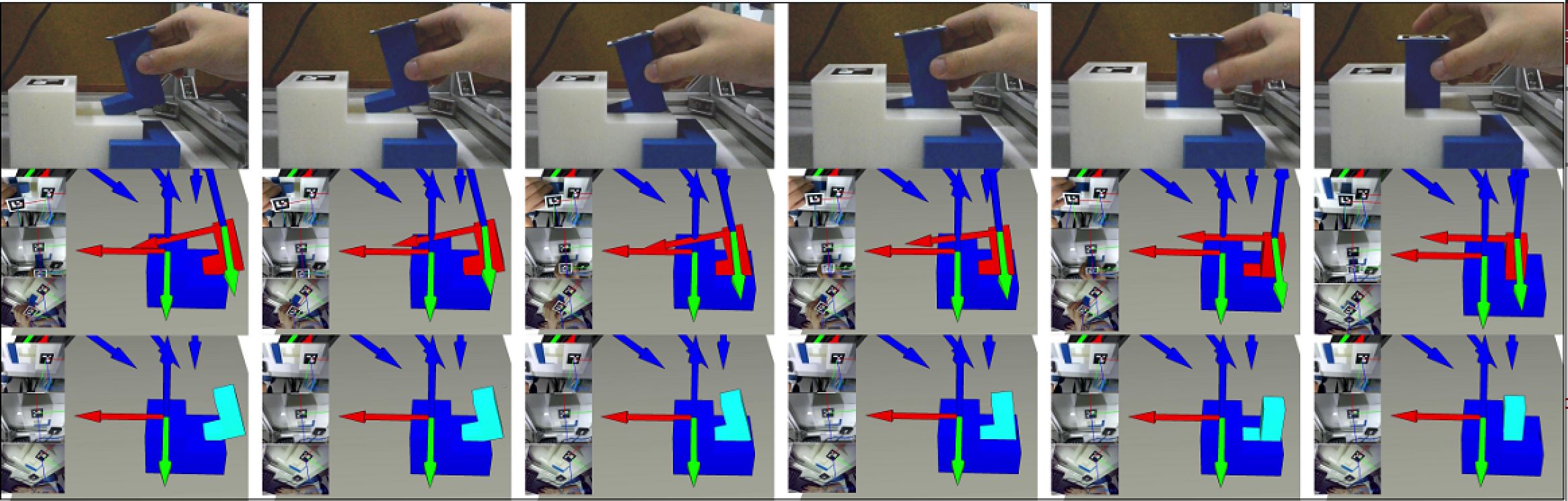}
            (a)
            \includegraphics[width=17cm]{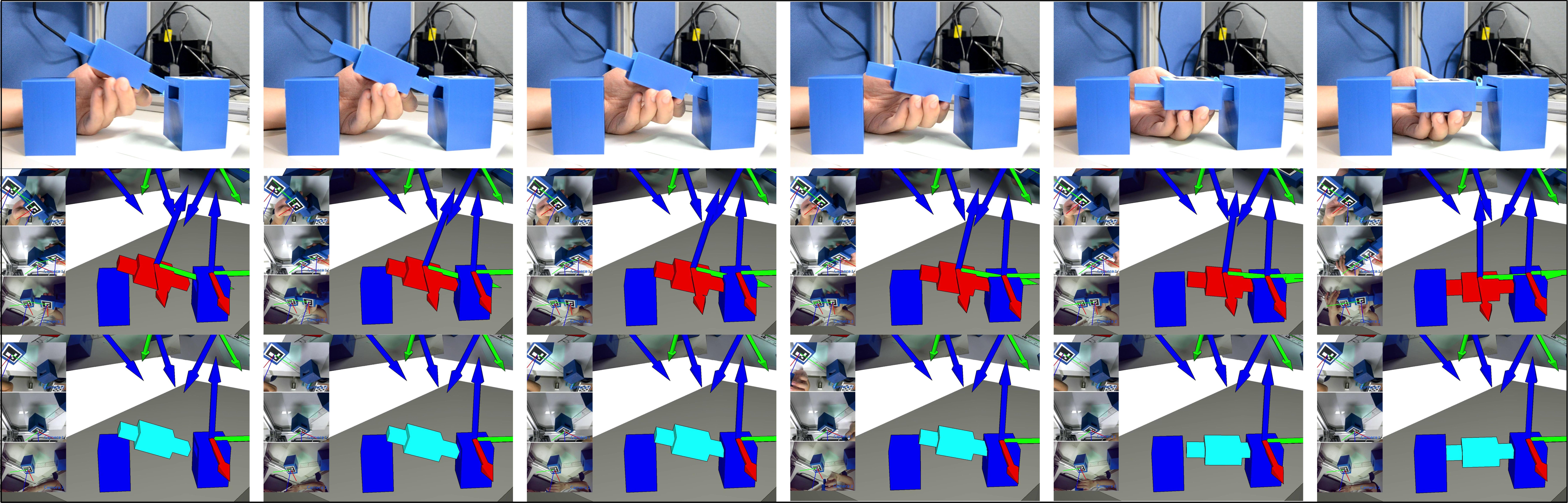}
            (b)
            \caption{These two figures are the human demonstration poses and corresponding poses captured in the virtual scene of (a) \textbf{L} insertion and (b) tenon insertion task. Human demonstration poses are shown in the top row. The middle row shows the tracking process through the vision system, and the bottom rows displays the final captured poses corresponding to human demonstration poses in each figure.}
            \label{fig:demonstration}
        \end{figure*}
In our experiment, we tested our method on two different tasks: the L-shaped object insertion and the tenon insertion. The tenon insertion is also a common assembly problem similar to the \textbf{L} insertion. and the target is inserting the tenon into the holes of the two mortises respectively.

\begin{figure}[t]
            \centering
            \includegraphics[scale=0.21]{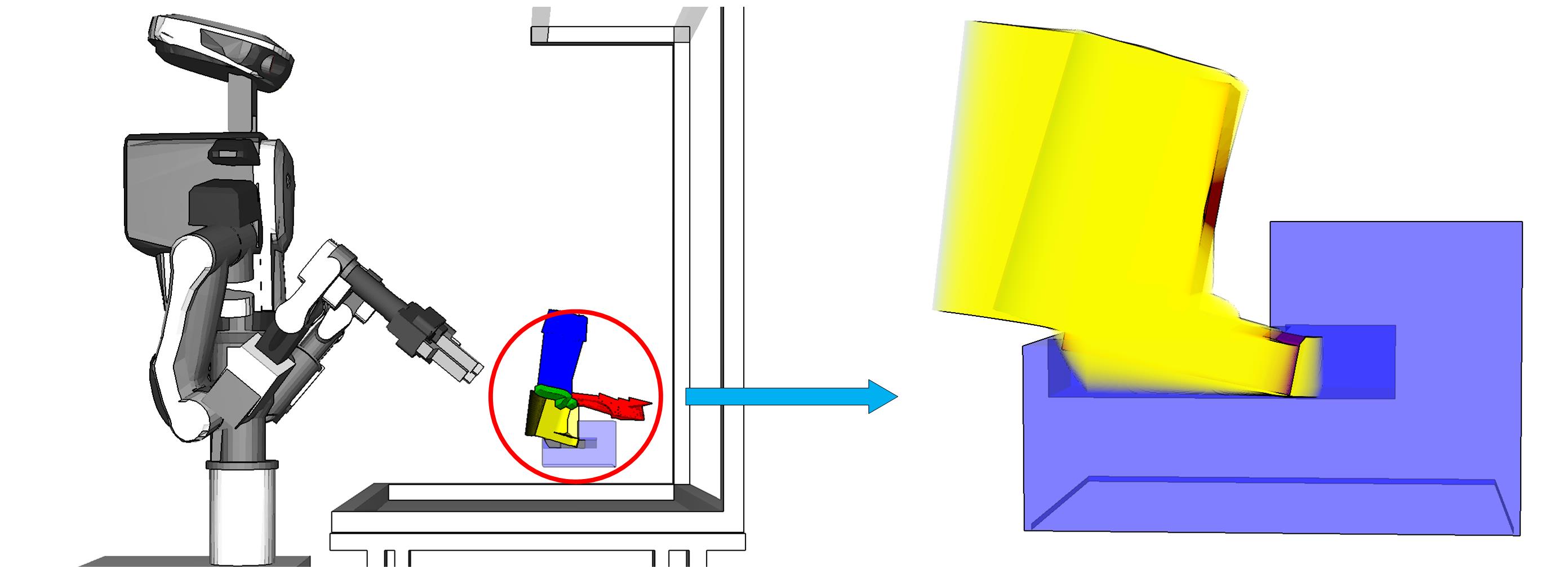}
            (a)
            \includegraphics[scale=0.30]{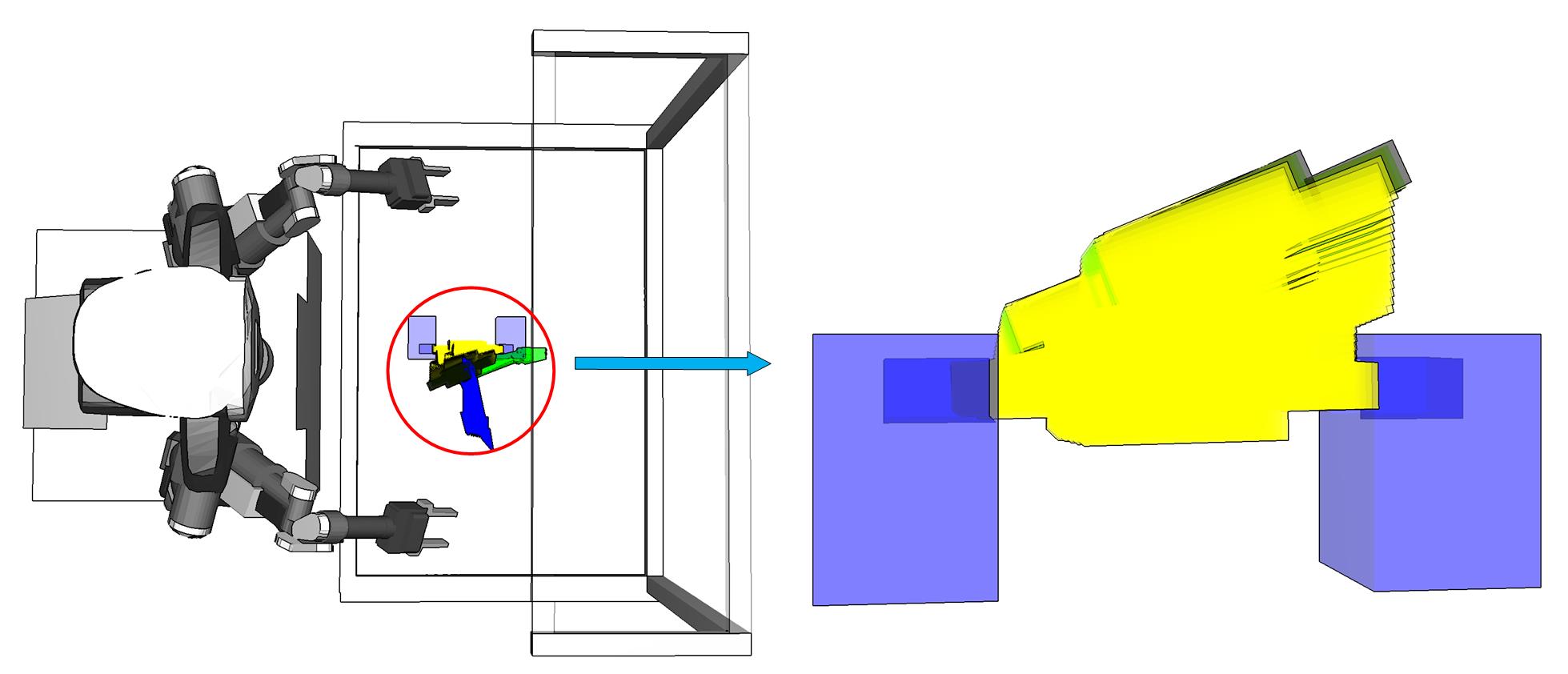}
            (b)
            \caption{The collision-free path of (a) \textbf{L} insertion and (b) tenon insertion generated by motion planning using demonstration.}
            \label{fig:freepath}
        \end{figure}

We managed to capture some key poses of the manipulated object with the vision system. Fig.\ref{fig:demonstration} shows key poses demonstrated by human and corresponding virtual scene obtained through the vision system on the screen. With the demonstration
poses, motion planning was implemented by the planner, given the initial pose and the final pose in Fig.\ref{fig:initfinpose}. Table \ref{table:planningresult} shows the results of motion planning. The time threshold $t_e$ is set to 5 seconds, which determines if the planning is time-resuming. Due to the feature of our planning strategy and algorithm, the results show differences both on time and used demonstration poses. However, it is evident that generally involving more demonstration poses brings better planning results and relatively less time-cost. The eventual obtained collision-free motion path is shown in Fig.\ref{fig:freepath}.

\renewcommand{\arraystretch}{1.5}
\setlength{\tabcolsep}{0.5pt}
\begin{table}[b]
\centering
    \caption{Motion Planning Results. The planning finishes when a path found in Less Than $t_e$=5 sec. }
    \begin{tabularx}{8cm}{p{2cm}<{\centering} p{1.6cm}<{\centering} p{1.2cm}<{\centering} p{1.6cm}<{\centering} X<{\centering}}
    \hline\hline
    \multirow{2}{*}{Task Scenario} & Candidate & \multirow{2}{*}{Trial} & Used & \multirow{2}{*}{Time (s)} \\
                                   & \multicolumn{1}{l}{$P^D$ Number} &       & \multicolumn{1}{l}{$P^D$ Number} &     \\
    \hline
    \multirow{5}{*}{\textbf{L }insertion} & \multirow{5}{*}{6} & \textbf{1}     & 2              & 1.510    \\
                                 &                     & \textbf{2}     & 3              & 1.714    \\
                                 &                     & \textbf{3}     & 3              & 1.431    \\
                                 &                     & \textbf{4}     & 2              & 3.034    \\
                                 &                     & \textbf{5}     & 2              & 2.172    \\
    \hline
    \multirow{5}{*}{Tenon insertion} & \multirow{5}{*}{11} & \textbf{1}     & 4              & 2.132    \\
                                 &                     & \textbf{2}     & 4              & 2.655    \\
                                 &                     & \textbf{3}     & 2              & 2.558    \\
                                 &                     & \textbf{4}     & 5              & 1.567    \\
                                 &                     & \textbf{5}     & 5              & 1.256    \\
    % Task & Candidate  & Used $P^D$ &\multicolumn{5}{c}{Time (s)} \\
    % \cline{4-8}    
    % Scenario & $P^D$ Number & Number & Test 1 & Test 2 & Test 3 & Test 4 & Test 5 \\
    % \hline
    % \textbf{L} insertion & 6 & 0 & \textit{NP} & \textit{NP} & \textit{NP} & \textit{NP} & \textit{NP} \\
    %                      &   & 1 & \textit{NP} & \textit{NP} & \textit{NP} & \textit{NP} & \textit{NP} \\
    %                      &   & 2 & 1.510 & \textit{NP} & \textit{NP} & 3.034 & 2.172 \\
    %                      &   & 3 & - & 1.714 & 1.431 & - & - \\
    % \hline
    % Tenon insertion & 11 & 0 & \textit{NP} & \textit{NP} & \textit{NP} & \textit{NP} & \textit{NP} \\
    %                 &    & 1 & \textit{NP} & \textit{NP} & \textit{NP} & \textit{NP} & \textit{NP} \\
    %                 &    & 2 & \textit{NP} & \textit{NP} & 2.558 & \textit{NP} & \textit{NP} \\
    %                 &    & 3 & \textit{NP} & \textit{NP} & - & \textit{NP} & \textit{NP} \\
    %                 &    & 4 & 2.132 & 2.655 & - & \textit{NP} & \textit{NP} \\
    %                 &    & 5 & - & - & - & 1.567 & 1.256 \\
    \hline\hline
    \end{tabularx}
    \label{table:planningresult}
\end{table}

\subsection{Actual Robot Experiments}
In order to validate the effectiveness of the proposed motion planning method, we have executed actual robot experiments on the Nextage OPEN humanoid robot. The robot control is realized through the ROS APIs of Nextage OPEN\footnote{http://wiki.ros.org/rtmros\_nextage/Tutorials} in PyManipulator.

    \begin{figure}[t]
                \centering
                \includegraphics[scale=0.25]{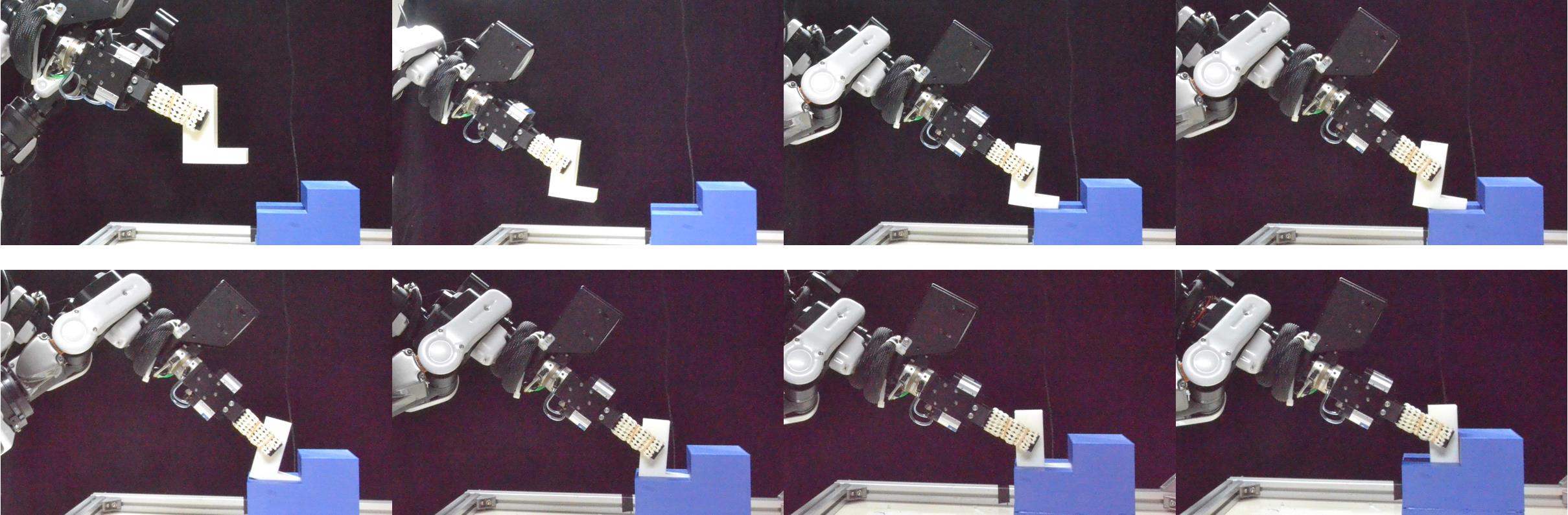}
                (a)
                \includegraphics[scale=0.5]{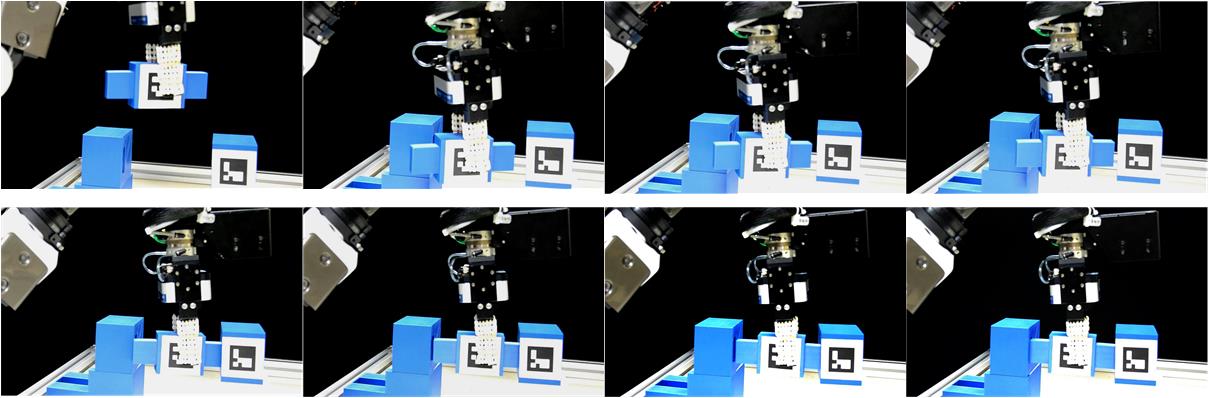}
                (b)
                \caption{Process of (a) the L-shaped object insertion and (b) the tenon insertion experiment}
                \label{fig:ex}
            \end{figure}
            
The two experiments were the \textbf{L} insertion experiment and the tenon insertion experiment.  Fig.\ref{fig:ex}(a) and (b) shows the experiment results of \textbf{L} insertion Trial 3 and tenon insertion Trial 5 respectively. The process is from the top-left picture to the bottom-right one.

For each scenario, 5 trials were conducted, in which the \textbf{L} insertion succeeded 4 times and the tenon insertion 3 times. The failure attempts should be ascribed to the errors of vision system resulting from the cameras and surrounding noise, inaccuracies of collision detection in motion planning, and the measurement error in experiments.

\section{Conclusions and Future Work}
This paper proposes an effective motion planning method making use of demonstration for dealing with complicated motions in assembly process. The effectiveness and feasibility of the method was instructed through the use of convenient AR marker detection, the high stability and precision of demonstration poses capture, and the fast planning speed.

It is proved that our approach can handle the narrow-space insertion problem well in this research. However, the practicality of the approach dealing with other kinds of skillful motions has not been verified and evaluated yet. In future, we will possibly apply this method to more sophisticated assembly process combining several kinds of skillful motions , where the assembly sequence may also be considered.

\addtolength{\textheight}{-12cm}   % This command serves to balance the column lengths

%%%%%%%%%%%%%%%%%%%%%%%%%%%%%%%%%%%%%%%%%%%%%%%%%%%%%%%%%%%%%%%%%%%%%%%%%%%%%%%%

%%%%%%%%%%%%%%%%%%%%%%%%%%%%%%%%%%%%%%%%%%%%%%%%%%%%%%%%%%%%%%%%%%%%%%%%%%%%%%%%

%%%%%%%%%%%%%%%%%%%%%%%%%%%%%%%%%%%%%%%%%%%%%%%%%%%%%%%%%%%%%%%%%%%%%%%%%%%%%%%%
\section*{ACKNOWLEDGMENT}

This paper is based on results obtained from a project commissioned by the New Energy and Industrial Technology Development Organization (NEDO). 

The first author would like to acknowledge the financial supports from the China Scholarship Council Postgraduate Scholarship Grant 201806120019.

%%%%%%%%%%%%%%%%%%%%%%%%%%%%%%%%%%%%%%%%%%%%%%%%%%%%%%%%%%%%%%%%%%%%%%%%%%%%%%%%
\bibliographystyle{unsrt}
\bibliography{references.bib}
\end{document}